\pdfoutput=1
\documentclass[11pt]{article}

\usepackage[]{acl}

\usepackage{times}
\usepackage{latexsym}

\usepackage[T1]{fontenc}
\usepackage[utf8]{inputenc}

\usepackage{microtype}

\usepackage{inconsolata}

\usepackage{bm}
\usepackage{adjustbox}
\usepackage{amsmath}
\usepackage{setspace}
\usepackage{graphicx}
\usepackage{amssymb}
\usepackage{dsfont}
\usepackage{multirow}
\usepackage{booktabs}
\usepackage{array}
\usepackage{pgfplots}
\usepgfplotslibrary{fillbetween}
\usetikzlibrary{patterns}
\usepackage{subcaption}
\usepackage{ragged2e}
\usepackage{mathtools}
\usepackage{xcolor}
\usepackage{framed}

\title{Protecting Privacy Through Approximating Optimal Parameters for Sequence Unlearning in Language Models}

\author{Dohyun Lee\textsuperscript{1*}\hspace{0.5cm} Daniel Rim\textsuperscript{2*}\hspace{0.5cm} Minseok Choi\textsuperscript{1}\hspace{0.5cm} Jaegul Choo\textsuperscript{1} \\
\textsuperscript{1}KAIST AI, \textsuperscript{2}Hyundai Motor Company \\
\texttt{\{aiclaudev, minseok.choi, jchoo\}@kaist.ac.kr} \\
\texttt{\{drim\}@hyundai.com}}

\begin{document}
\maketitle
\def\thefootnote{*}\footnotetext{\hspace{0.1cm} Equal contribution}\def\thefootnote{\arabic{footnote}}
\begin{abstract}
Although language models (LMs) demonstrate exceptional capabilities on various tasks, they are potentially vulnerable to extraction attacks, which represent a significant privacy risk. 
To mitigate the privacy concerns of LMs, machine unlearning has emerged as an important research area, which is utilized to induce the LM to selectively forget about some of its training data. 
While completely retraining the model will guarantee successful unlearning and privacy assurance, it is impractical for LMs, as it would be time-consuming and resource-intensive.
Prior works efficiently unlearn the target token sequences, but upon subsequent iterations, the LM displays significant degradation in performance.
In this work, we propose \textbf{P}rivacy Protection via \textbf{O}ptimal \textbf{P}arameters (POP), a novel unlearning method that effectively forgets the target token sequences from the pretrained LM by applying optimal gradient updates to the parameters.
Inspired by the gradient derivation of complete retraining, we approximate the optimal training objective that successfully unlearns the target sequence while retaining the knowledge from the rest of the training data.
Experimental results demonstrate that POP exhibits remarkable retention performance post-unlearning across 9 classification and 4 dialogue benchmarks, outperforming the state-of-the-art by a large margin.
Furthermore, we introduce Remnant Memorization Accuracy that quantifies privacy risks based on token likelihood and validate its effectiveness through both qualitative and quantitative analyses.

\end{abstract}

\begin{figure}[t]
\centering
\includegraphics[width=\linewidth]{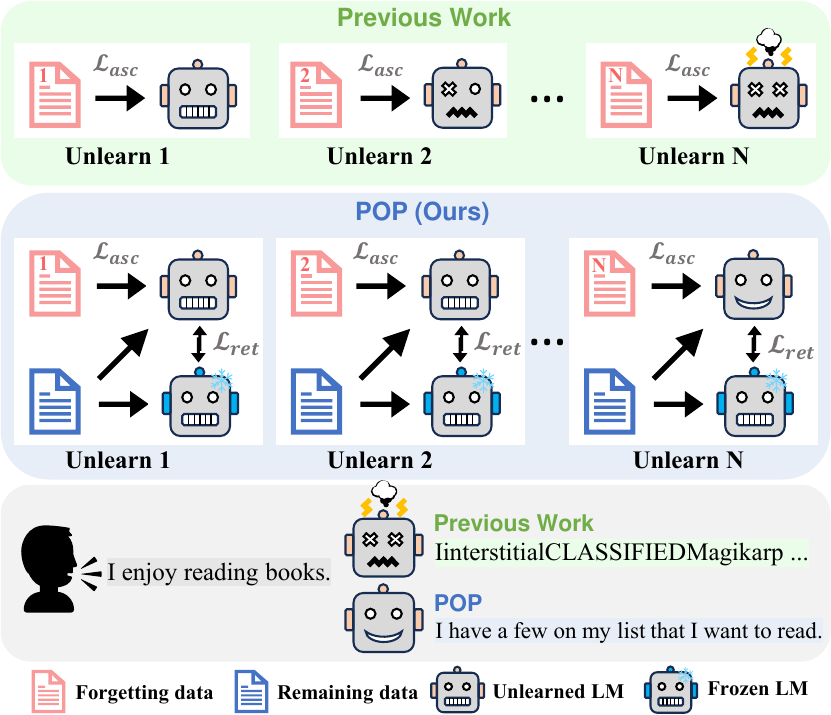}
\caption{
\textbf{Our proposed method.}
$\mathcal{L}_{asc}$ is the gradient ascent loss for unlearning the target data. 
If utilized alone, significant performance degradation occurs. 
By applying both retain loss $\mathcal{L}_{ret}$ and $\mathcal{L}_{asc}$, our method unlearns the target data \textit{and} retains the LM performance.
For example, after applying unlearning in succession, previous work demonstrates catastrophic degradation, while POP demonstrates successful retention. 
Our approach is detailed in Section~\ref{sec:Methodology}.}
\label{fig:example}
\vspace{-0.3cm}
\end{figure}

\section{Introduction}
Language models (LMs) pretrained on a substantial amount of text have demonstrated remarkable performance on various tasks. 
One of the most important factors in improving performance is training on larger datasets, often containing more than trillions of tokens in the latest models.
The datasets used to train such models, however, inevitably contain private information, as it is impossible to check all tokens for privacy concerns.  
Machine learning models are well-known for being vulnerable to manipulations that can expose the training data, potentially generating exact strings from the training data~\cite{carlini2019secret, carlini2021extracting}. 
Additionally, it has been reported that extracting exact training data becomes easier as models scale to larger sizes~\cite{carlini2022quantifying}. 
With many LMs publicly available~\cite{zhao2023survey}, the importance of managing the inherent privacy risks in such models has also increased. 
Moreover, all practitioners are required to delete personal information from machine learning models when requested, to comply with the ``Right To Be Forgotten (RTBF)''~\cite{hoofnagle2019gdpr} from the European Union's General Data Protection Regulation agreement~\cite{voigt2017gdpr} and the United States California Consumer Privacy Act~\cite{pardau2018california}. 
To mitigate the potential data leakage and comply with privacy regulations, machine unlearning has emerged as an important research area. 

Previous machine unlearning approaches attempted to achieve exact unlearning by removing all private information from the training data, or designed algorithms to ensure differential privacy (DP)~\cite{anil2021large}.
Some have proposed changes to the training process to make unlearning easier for the pretrained model~\cite{thudi2022unrolling}.  
These efforts require re-training of LMs every time an individual practices one's RTBF, which is extremely expensive and time-consuming.
Although the complete re-training of LMs would be optimal for machine unlearning, the cost of doing so is too severe, making such approaches impractical. 
Others proposed approximate unlearning of target token sequence, applying only a few parameter updates to the pretrained LMs~\cite{jang2023knowledgeunlearning}, or utilizing reinforcement learning feedback loop via proximal policy optimization to unlearn the token sequences~\cite{kassem-etal-2023-preserving}. 
\citet{jang2023knowledgeunlearning} assert that a simple gradient ascent on the target token sequences can be effective at forgetting them.
This method is not optimal, as gradient ascent only applies a \textit{portion} of the optimal gradient updates to the parameters. 
As shown in Fig.~\ref{fig:example}, adherence to multiple unlearning requests results in accumulation of errors from inadequate approximations, ultimately accumulating to a significant amount. 
While it may successfully unlearn a few instances in a single batch, the degradation in performance will make the LM useless after unlearning multiple sequences.
As ensuring the retention of LM performance is just as important as unlearning the target token sequences, any method that cannot guarantee unlearning \textit{and} retention, after multiple requests, is not a viable machine unlearning solution.
\citet{kassem-etal-2023-preserving} demonstrated better retention of language model capabilities in various NLP benchmarks, but their method requires all token sequences that come before the target token sequence in the training data to unlearn the target token sequence. As there can be multiple token sequences that come before a target token sequence, their method is extremely difficult to apply in real-world applications.

In this paper, we propose \textbf{P}rivacy Protection via \textbf{O}ptimal \textbf{P}arameters (POP), which applies the optimal gradient updates for sequence unlearning. 
The gold standard for machine unlearning is a complete retraining from scratch, after removing the target token sequences from the training data. 
Without committing excessive approximations, POP attempts to emulate the gold standard, updating the parameters as if they were never trained on the target token sequence.  
After carefully examining the overall gradient updates of the training process, we identify the optimal parameter updates for machine unlearning.
Based on our findings, we formalize our solution, which utilizes the pretrained weights, the target token sequence, and the remaining data to achieve inexpensive and optimal machine unlearning. 
As shown in Fig.~\ref{fig:example}, POP successfully unlearns the target sequence and ensures the retention of general LM performance post-unlearning, even in a sequential unlearning context where the model applies unlearning requests in succession. 
Moreover, POP does not require any token prefixes from the training data to unlearn token sequences, rendering it a more viable choice in real-world settings.

We also present Remnant Memorization Accuracy (RMA), a novel metric for quantifying privacy risks. 
Compared to other sequence unlearning metrics, RMA is the most strict and provides the most robust privacy protection, as it considers the \textit{probabilities} of tokens within the target sequences. 
When utilized in an unlearning context, RMA can be used as a guideline to determine when unlearning is completed. 
As it would be unnecessary to excessively unlearn the target sequence from the model, setting an appropriate threshold for unlearning is important. 
We perform experiments by setting empirical thresholds for each unlearning metric and demonstrate RMA's superiority in providing the strongest privacy protection. 

Overall, our contributions are threefold:
\vspace{-2mm}
\begin{itemize}
    \item We present POP, a robust knowledge unlearning method that successfully unlearns a target sequence while retaining the general performance of the LM.
    \vspace{-2mm}
    \item We demonstrate POP's superior performance in both the batch and sequential unlearning processes through quantitative and qualitative analyses.  
    \vspace{-2mm}
    \item We propose RMA, a novel metric for quantifying privacy risks, and demonstrate its strength in providing robust privacy guarantees.
\end{itemize}

\setlength{\abovedisplayskip}{7pt}
\setlength{\belowdisplayskip}{7pt}

\section{Related Work}
\paragraph{Data Preprocessing}

This approach aims to achieve exact unlearning by removing the target sequences from training data through preprocessing methods.
This can effectively mitigate privacy risks for sequences that follow easily identifiable formats, such as phone numbers, email addresses, and more~\cite{aura2006scanning, dernoncourt2017identification, lison2021anonymisation}.
Private information, however, is context-dependent~\cite{brown2022does}, making it impossible to completely remove all private data. 
Another method that is applied prior to training is data deduplication~\cite{kandpal2022deduplicating}, which showed improved robustness against data extraction attacks by removing duplicate data from the pretraining corpus. 
Although this may be effective at mitigating overall privacy risks, it cannot be utilized in a targeted manner for unlearning a specific target token sequence. 

\paragraph{Differential Privacy}
DP preserving methods look to prevent memorization of individual training examples~\cite{dwork2006calibrating, dwork2006differential, abadi2016deep}.
Although such methods have been effective in fine-tuning LMs~\cite{yu2021differentially, li2021large}, pretraining LMs with DP significantly reduces performance, requires expensive computations, and converges very slowly~\cite{anil2021large}. 
Furthermore, as it is impossible to define privacy boundaries for natural language~\cite{brown2022does}, DP methods are inherently not applicable for target sequence unlearning. 

\paragraph{Knowledge Editing}
Knowledge editing methods modify LMs to achieve a diverse set of objectives. 
Some apply various transformations to the neural representations to identify and remove specific concepts~\cite{ravfogel2022adversarial, ravfogel2022linear, belrose2023leace}. 
Some apply other methods to maintain the relevancy of the LMs, efficiently updating the underlying knowledge without degrading their performance~\cite{yao2023editing}. 
Although these methods alter the pretrained LM for their respective goals, none are designed for the task of unlearning specific token sequences. 

\paragraph{Sequence Unlearning}
For unlearning specific token sequences, \citet{jang2023knowledgeunlearning} proposed a simple gradient-based solution in reducing the generation likelihood of forgetting token sequences. 
Although the proposed solution can approximately remove a target token sequence, it also suffers from a large degradation in overall language modeling performance. 
This downside is even more evident when unlearning multiple sequences in succession, making it impractical for real-world use. 
Our method not only effectively trains the LMs to forget the target sequence, but also mitigates the potential problems from approximation of the gradients. 

More recently, \citet{kassem-etal-2023-preserving} presented DeMem, which utilizes a reinforcement learning feedback loop via proximal policy optimization to unlearn token sequences that follow the given prefix sequences. 
Although DeMem achieves sequence unlearning, it is fundamentally different from ours as their goal is to mitigate memorization by altering the token sequences that follow the given prefix sequences. 
In a real-world setting with multiple RTBF requests, however, defining the correct set of prefixes for a target token sequence will be difficult, and missing a prefix could present privacy concerns. 
An ideal unlearning solution should remove token sequences without relying on identifying all possible prefix sequences. 
POP provide a more robust unlearning solution, by eliminating the generation likelihood of any token sequences.

\section{Methodology}
\label{sec:Methodology}

\subsection{Problem Definition}
\label{sec:task_formalization}
Given $i$-th sequence of tokens $\mathbf{x}_i=(x_1, \dots, x_T)$ in the pretraining dataset $\mathcal{D} = \{\mathbf{x}_1, \dots, \mathbf{x}_N\}$, causal language modeling minimizes the negative log-likelihood loss: 
\vspace{-3mm}
\begin{equation}
\mathcal{L}(\mathbf{x}_i;\theta)=-\sum_{t=1}^T\text{log}(p_{\theta}(x_t|x_{<t})).
\label{eq1}
\vspace{-1mm}
\end{equation}
Assuming that the update occurred for each sequence, and without considering the learning rate, we define the update step as
\begin{equation}
\theta_j = \theta_{j-1}-\nabla_{\theta}\mathcal{L}(\mathbf{x}_j;\theta_{j-1}),
\end{equation}
where $\theta_j$ denotes the parameters which is updated for each sequence on $\{\mathbf{x}_1, \dots, \mathbf{x}_j\}$.
Notably, the pretrained model $\theta_{\text{ptr}}$ is equal to $\theta_{N}$, as both are trained on $N$ token sequences.
Subsequently, our unlearning objective is to approximate the optimal parameters achievable from complete retraining, i.e., $\theta_{\text{rtr}}$, from the pretrained model $\theta_{\text{ptr}}$.
Concretely, $\theta_{\text{ptr}}$ refers to the parameters before unlearning the target sequence $\mathbf{x}^F \in \mathcal{D}^F$, where $\mathcal{D}^F \subset \mathcal{D}$ contains the target sequence, and $\theta_{\text{rtr}}$ denotes the optimal parameters obtained from retraining on the remaining data $\mathcal{D}^R = \mathcal{D} \setminus \mathcal{D}^F$.

\subsection{POP}
\label{sec:pop}

In this section, we elaborate on the details of POP and its derivations for the optimal parameter updates for sequence unlearning.

\paragraph{Approximation of $\theta_{\text{rtr}}$} 
Suppose that the arbitrary sequence $\mathbf{x}_n \in \mathcal{D}$ for $1 \leq n \leq N$ is the target sequence $\mathbf{x}^F$.
Then, $\theta_{\text{rtr}}$ is updated on $\mathcal{D}$ except for $\mathbf{x}_n$ from the randomly initialized parameters $\theta_0$:
\vspace{-1mm}
\begin{equation} 
    \begin{aligned}
        \theta_{\text{ptr}} = \theta_{0} - \sum_{i=1}^N\nabla_{\theta}\mathcal{L}(\mathbf{x}_i;\theta_{i-1}),
    \end{aligned}
\label{eq3}
\end{equation}
\vspace{-6mm}
\begin{equation}
    \text{\normalsize$\theta_{\text{rtr}}\!\!=\!\theta_{0}$}\! - \!\!\text{\normalsize$\sum_{i=1}^{{n-1}}\nabla_{\theta}\mathcal{L}(\mathbf{x}_i;\theta_{i-1})$}\! -\!\!\!\!\text{\normalsize$\sum_{i=n+1}^{N}\!\!\!\nabla_{\theta}\mathcal{L}(\mathbf{x}_i;\theta^*_{i-1})$},
\end{equation}

where $\theta^*_j$ refers to the parameters trained on $\{\mathbf{x}_1, \cdots, \mathbf{x}_j\}$ without the target sequence $\mathbf{x}_n$ for $n \leq j$. 
In other words, $\theta_{\text{rtr}}$ is equal to $\theta_{N}^*$, since it is trained on $\{\mathbf{x}_1, \dots, \mathbf{x}_N\}$ except for $\mathbf{x}_n$.
By leveraging the equations above, we can derive the equation where $\theta_{\text{rtr}}$ is represented by $\theta_{\text{ptr}}$:
\vspace{-1mm}
\begin{gather}
\theta_{\text{rtr}}=\theta_{\text{ptr}}+\nabla_{\theta}\mathcal{L}(\mathbf{x}_n;\theta_{n-1})+S, \label{eq:4} \\
S=\sum_{i=n+1}^N\nabla_{\theta}\mathcal{L}(\mathbf{x}_i;\theta_{i-1})-\nabla_{\theta}\mathcal{L}(\mathbf{x}_i;\theta^*_{i-1}). \label{eq:5}
\end{gather}
\paragraph{Derivation of a Tractable Solution}
Although the derived equation above is reasonable, we cannot compute the $\sum$ \ in Equation~\ref{eq:5} because $\theta$s during training are intractable.
To address this, we constrain $\mathbf{N \approx n+1}$, where we suppose the target sequence $\mathbf{x}_n$ is trained just before the last sequence:
\begin{gather}
S=\nabla_{\theta}\mathcal{L}(\mathbf{x}_{n+1};\theta_{n})-\nabla_{\theta}\mathcal{L}(\mathbf{x}_{n+1};\theta_{n-1}), \label{eq:6}
\end{gather}
where $\mathbf{x}_{n+1}$ refers to remaining data $\mathbf{x}^{R} \! \in \! \mathcal{D}^R$ without the target sequence $\mathbf{x}_{n}( \! = \! \mathbf{x}^{F})$, and we can say that $\theta_{n}$ has more knowledge of the target sequence than $\theta_{n-1}$ does. 

\paragraph{Iterative Update Equation}
Using Equations~\ref{eq:4} and~\ref{eq:6}, we initialize $\theta_{n-1}$ with $\theta_{\text{ptr}}$, which is iteratively updated to unlearn the target sequence $\mathbf{x}^F$.
To assure the relationship between $\theta_{n}$ and $\theta_{n-1}$, we fix $\theta_n$ as $\theta_{\text{ptr}}$, where the parameters remain frozen during unlearning.
Then, the iterative update equation for unlearning the target sequence is
\begin{gather}
\theta:=\theta+\nabla_{\theta}\mathcal{L}(\mathbf{x}^{F};\theta)+S, \\
S=\nabla_{\theta}\mathcal{L}(\mathbf{x}^{R};\theta_{\text{ptr}})-\nabla_{\theta}\mathcal{L}(\mathbf{x}^{R};\theta),
\end{gather}
where $\theta$ is trainable parameters initialized with $\theta_{\text{ptr}}$, and is unlearned until convergence to $\theta_{\text{rtr}}$.
% , which is initialized with $\theta_{\text{ptr}}$ and will be updated 
\paragraph{From Gradients to Loss Terms}
% In the experiment, we cast aspects of gradient as loss and use the following loss terms for ease of implementation:
For training, we use the following losses corresponding to the derived gradient terms:
% \begin{gather}
% \mathcal{L}_{\text{pop}} = \mathcal{L}_{\text{asc}} + \lambda \mathcal{L}_{\text{ret}} \\
% \mathcal{L}_{asc} = \mathbb{E}_\mathit{D^F}[\text{log}(p_{\theta}(\mathbf{x}))] \\
% \mathcal{L}_{\text{ret}} = \mathbb{E}_\mathit{D^R}[\text{log}(p_{\theta^F}(\mathbf{x}))] - \mathbb{E}_\mathit{D^R}[\text{log}(p_{\theta}(\mathbf{x}))] 
% \end{gather}
\begin{gather}
\mathcal{L}_{\text{asc}} = \mathbb{E}_{\mathcal{D}^F}[\log(p_{\theta}(\mathbf{x}))], \label{eq:9} \\
\mathcal{L}_{\text{ret}} = \mathbb{E}_{\mathcal{D}^R}[\log(p_{\theta_{\text{ptr}}}(\mathbf{x})) - \log(p_{\theta}(\mathbf{x}))], \label{eq:10} 
% \mathcal{L}_{\text{ret}} = \mathbb{E}_\mathit{D^R}[\text{log}(p_{\theta^F}(\mathbf{x})) - \text{log}(p_{\theta}(\mathbf{x}))] \label{eq:12}
\end{gather} 
where $\mathcal{L}_{\text{asc}}$ refers to the loss for unlearning the target sequence $\mathbf{x}^F \in \mathcal{D}^F$, while $\mathcal{L}_{ret}$ denotes the loss associated with retaining the remaining data $\mathbf{x}^R \in \mathcal{D}^R$ performance.
Putting everything together, the overall training objective for sequence unlearning is minimizing the following loss: 
\begin{equation}
\mathcal{L}_{\text{pop}} = \mathcal{L}_{\text{asc}} + \lambda \mathcal{L}_{\text{ret}}, \label{eq:11}
\end{equation}
where $\lambda$ is a loss scaling hyperparameter.
In $\mathcal{L}_{ret}$, the first term is ignored by the optimization, even though it contains the initial state of the pretrained LM.
Since this leads to underutilization of the pretrained LM for retaining the remaining data, we use the probability distribution over the vocabulary of the pretrained LM as the soft labels. 
This is quite intuitive, as the objective of POP is to unlearn the target token sequence \textit{without} deviating too much from the initial state of the pretrained LM.

\subsection{Remnant Memorization Accuracy}
\label{sec:unlearning_metric}
% \vspace{-3mm}
Given a sequence of tokens $\mathbf{x}=(x_1, \dots, x_T)$, previous studies have proposed metrics to assess ``how well a model remembers a specific sequence of tokens'', and unlearning can be achieved by decreasing the value of these metrics for the forgetting data.
\citet{tirumala2022memorization} and \citet{jang2023knowledgeunlearning} suggested Memorization Accuracy (MA) and Extraction Likelihood (EL), respectively:
\begin{gather}
\text{MA}=\frac{\sum_{t=1}^{T-1} \mathds{1}\{\text{argmax}(p_\theta(\cdot|x_{<t}))=x_t\}}{T-1}
\end{gather}
\begin{gather}
\text{EL}_n=\frac{\sum_{t=1}^{T-n}\text{OVERLAP}_n(f_\theta({x_{<t}}), x_{\geq t})}{T-n} \\
\text{OVERLAP}_n(a, b)=\frac{\sum_{c \in ng(a)} \mathds{1} \{c \in ng(b)\}}{|ng(a)|}, \notag
\label{eq:EL}
\end{gather}
where $\mathit{ng(\cdot)}$ in EL represents the list of n-grams in the given sequence, and $\mathit{f_\theta({x_{<t}})}$ represents the output sequence from the LM.
As unlearning metrics are often utilized to determine the thresholds for unlearning, thereby setting the stopping point of the unlearning process, it is important that they accurately portray the privacy risk of LM post-unlearning. 
MA and EL, however, disregard the probabilities of tokens within the sequence.
In other words, they do not consider the situation where the target token has the second highest probability in the probability distribution for the next token prediction.
When these metrics are used to determine the stopping point of the unlearning process, the resulting LM can be vulnerable to various attacks that could extract the target token through sampling methods.

\begin{figure}[t]
\centering
\includegraphics[width=\linewidth]{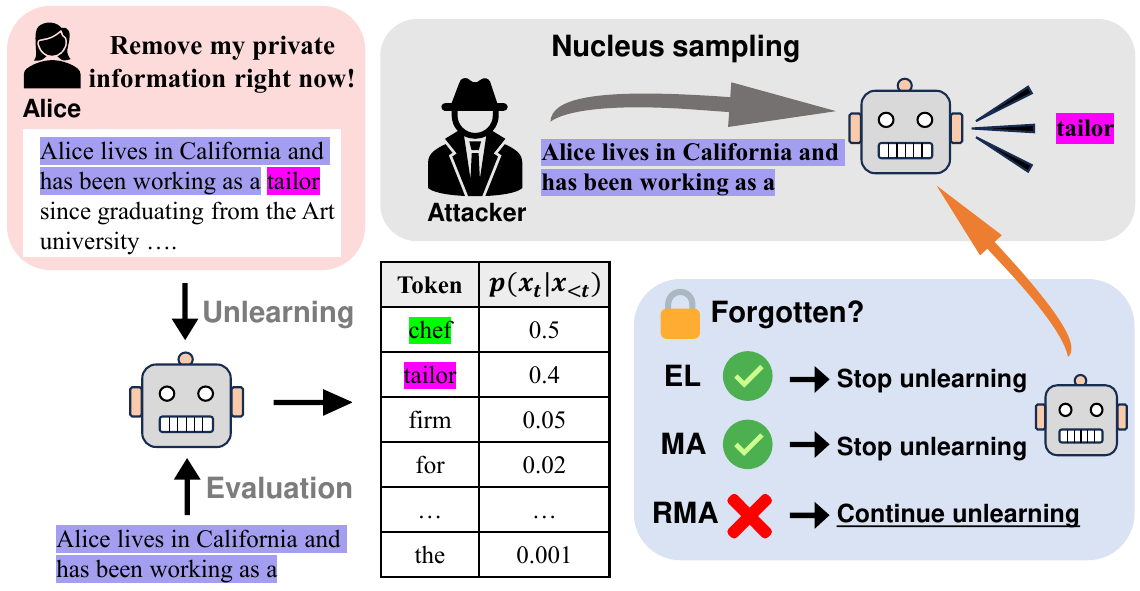}
\caption{
\textbf{Privacy Protection of RMA.} Compared to other metrics, RMA considers the token probabilities to better represent the inherent privacy risk, and provides the strongest privacy protection. 
} 
\label{fig:toyexample}
\end{figure}

To alleviate this limitation, we propose Remnant Memorization Accuracy (RMA):
\begin{gather}
\text{RMA}=\frac{\sum_{t=1}^{T-1} p_\theta(x_t|x_{<t})}{T-1}.
\end{gather}
Unlike other unlearning metrics, RMA considers the probabilities of tokens to better represent the privacy risk. 
Models unlearned until they satisfy the forgetting thresholds for RMA are significantly less likely to be vulnerable to extraction attacks.
When utilized individually, RMA is a more stringent unlearning metric, as it is more difficult to satisfy the forgetting threshold. 
Figure \ref{fig:toyexample} shows an example of how RMA can provide a stronger privacy protection compared to other unlearning metrics. 
The process for obtaining the forgetting thresholds is in Section \ref{sec:forgetthresholds}, and metric comparisons can be found in Section \ref{sec:metric_analysis}. 

\begin{table}[]
\centering
\small{
\begin{tabular}{@{}cl|ccc@{}}
\toprule
Model                & Size & EL\textsubscript{10} & MA & RMA \\ \midrule
\multirow{3}{*}{\raisebox{-2.0mm}{\hspace{1mm} OPT}} & 125M & 4.3      & 40.1    & 31.0    \\ \cmidrule(l){2-5} 
                     & 1.3B & 5.9      & 46.4    & 38.4    \\ \cmidrule(l){2-5} 
                     & 2.7B & 6.3      & 47.7    & 39.9    \\ \midrule \midrule
\multirow{3}{*}{\raisebox{-2.0mm}{\hspace{1mm} GPT-Neo}} & 125M & 6.3      & 48.7    & 41.5    \\ \cmidrule(l){2-5} 
                     & 1.3B & 7.9      & 54.2    & 48.1    \\ \cmidrule(l){2-5} 
                     & 2.7B & 8.5      & 55.5    & 49.6    \\ \bottomrule
\end{tabular}
}
\caption{Forgetting Thresholds}
\label{forgetting_threshold}
\end{table}

\section{Experimental Setup}
\label{sec:experiment}
\subsection{Baselines}
We experiment on two LMs for model sizes 125M, 1.3B, 2.7B: GPT-Neo LMs~\cite{black2021gpt} initially pretrained on the Pile~\cite{gao2020pile} corpus, and OPT LMs~\cite{zhang2022opt}, which are pretrained on a deduplicated version of the Pile, along with other corpora.  
We perform experiments with the following unlearning methods: 
\vspace{-2mm}
\begin{itemize}
    \item \textbf{UL}~\cite{jang2023knowledgeunlearning} decreases the log-likelihood of the target token sequences -- namely, only using $\mathcal{L}_{\text{asc}}$ in Equation~\ref{eq:11}.
    \vspace{-3mm}
    \item \textbf{POP\textit{\textsuperscript{$\flat$}}}~\cite{popb} utilizes $\mathcal{L}_{\text{asc}}$ and $\mathcal{L}_{\text{ret}}$ with the hard labels in Equation~\ref{eq:11}.
    \vspace{-3mm}
    \item \textbf{POP}, our main proposed method, utilizes $\mathcal{L}_{\text{asc}}$ and $\mathcal{L}_{\text{ret}}$ similarly to \textbf{POP\textit{\textsuperscript{$\flat$}}}, where $\mathcal{L}_{\text{ret}}$ uses the probability distribution over the vocabulary of the pretrained LM as the soft labels.
\end{itemize}
In Equation~\ref{eq:11}, we set the $\lambda$ as 1 for simplicity.

\subsection{Target Data Curation}
\label{datacuration}
We source the target sequence data from the Training Data Extraction Challenge\footnote{\url{https://github.com/google-research/lm-extraction-benchmark}}. 
This data consists of 15,000 examples, each not exceeding 200 tokens in length. 
In our experiments, we construct 19 target sequence datasets, each with 32 sequences.
Due to copyright issues, we randomly sample the remaining data from the uncopyrighted Pile corpus\footnote{\url{https://huggingface.co/datasets/monology/pile-uncopyrighted}}, without the target sequence. 

\begin{table*}[h]
\vspace{-1cm}
\centering
\small{
\begin{adjustbox}{width=0.8\textwidth}
\begin{tabular}{@{}cc|ccc|cc|c@{}}
\toprule
Model                     & Method  & EL\textsubscript{10} & MA  & RMA  & Classification (Acc) & Dialogue (F1)  & Epochs \\ \midrule
\multirow{4}{*}{\raisebox{-1.0mm}{\hspace{1mm} OPT-125M}} & Pretrained & 6.2  & 53.0 & 40.5 & 42.6           & 10.8       & -     \\ \cmidrule(l){2-8} 
                          & UL      & 2.7  & 29.8 & 28.7 & 32.9 \scriptsize{$\pm 0.37$}     & 1.9 \scriptsize{$\pm 0.47$}  & 8.4   \\
                          & POP\rlap{\textit{\textsuperscript{$\flat$}}}    & 3.5  & 29.8 & 22.8 & 37.0 \scriptsize{$\pm 1.18$}     & 4.1 \scriptsize{$\pm 1.39$}  & 8.4   \\
                          & POP     & 2.3  & 31.3 & 30.2 & \textbf{43.3} \scriptsize{$\pm 0.30$}     & \textbf{9.2} \scriptsize{$\pm 0.65$}  & 16.4  \\ \midrule
\multirow{4}{*}{\raisebox{-1.0mm}{\hspace{1mm} OPT-1.3B}} & Pretrained & 23.1 & 68.4 & 60.6 & 51.5           & 13.3       & -     \\ \cmidrule(l){2-8} 
                          & UL      & 2.7  & 32.0 & 30.9 & 36.2 \scriptsize{$\pm 1.74$}     & 1.8 \scriptsize{$\pm 1.47$}  & 5.6   \\
                          & POP\rlap{\textit{\textsuperscript{$\flat$}}}    & 2.1  & 38.4 & 34.3 & 42.4 \scriptsize{$\pm 0.62$}     & 5.5 \scriptsize{$\pm 0.57$}  & 6.2   \\
                          & POP     & 2.3  & 35.6 & 34.4 & \textbf{50.4} \scriptsize{$\pm 0.34$}     & \textbf{12.3} \scriptsize{$\pm 0.44$} & 7.8   \\ \midrule 
\multirow{4}{*}{\raisebox{-1.0mm}{\hspace{1mm} OPT-2.7B}} & Pretrained & 25.3 & 70.2 & 63.1 & 53.8           & 13.7       & -     \\ \cmidrule(l){2-8} 
                          & UL      & 2.7  & 34.1 & 33.4 & 37.0 \scriptsize{$\pm 2.36$}     & 1.2 \scriptsize{$\pm 1.65$}  & 6.2   \\
                          & POP\rlap{\textit{\textsuperscript{$\flat$}}}    & 3.2  & 41.7 & 37.6 & 42.1 \scriptsize{$\pm 2.24$}     & 7.0 \scriptsize{$\pm 0.42$}  & 8.8   \\
                          & POP     & 3.7  & 37.5 & 36.8 & \textbf{52.2} \scriptsize{$\pm 0.35$}     & \textbf{13.3} \scriptsize{$\pm 0.22$} & 10.6  \\ \midrule \midrule
\multirow{4}{*}{\raisebox{-1.0mm}{\hspace{1mm} Neo-125M}} & Pretrained & 36.1 & 77.9 & 71.1 & 43.5           & 10.0       & -     \\ \cmidrule(l){2-8} 
                          & UL      & 2.3  & 45.7 & 39.5 & 40.8 \scriptsize{$\pm 1.87$}     & 8.0 \scriptsize{$\pm 1.55$}  & 10.4  \\
                          & POP\rlap{\textit{\textsuperscript{$\flat$}}}    & 2.2  & 46.2 & 39.4 & 42.9 \scriptsize{$\pm 0.13$}     & 10.0 \scriptsize{$\pm 0.29$} & 14.6  \\
                          & POP     & 2.6  & 45.8 & 40.4 & \textbf{43.0} \scriptsize{$\pm 0.32$}     & \textbf{10.4} \scriptsize{$\pm 0.16$} & 13.2  \\ \midrule
\multirow{4}{*}{\raisebox{-1.0mm}{\hspace{1mm} Neo-1.3B}} & Pretrained & 66.0 & 92.1 & 88.3 & 49.7           & 12.3       & -     \\ \cmidrule(l){2-8} 
                          & UL      & 2.9  & 47.3 & 42.5 & 49.2 \scriptsize{$\pm 1.54$}     & 11.5 \scriptsize{$\pm 0.78$} & 5.4   \\
                          & POP\rlap{\textit{\textsuperscript{$\flat$}}}    & 2.8  & 48.3 & 43.9 & 48.3 \scriptsize{$\pm 0.31$}     & \textbf{12.1} \scriptsize{$\pm 0.16$} & 6.8   \\
                          & POP     & 3.2  & 48.8 & 44.4 & \textbf{49.5} \scriptsize{$\pm 0.34$}     & \textbf{12.1} \scriptsize{$\pm 0.19$} & 6.0   \\ \midrule
\multirow{4}{*}{{\raisebox{-1.0mm}{\hspace{1mm} Neo-2.7B}}} & Pretrained & 69.7 & 93.4 & 90.7 & 52.2           & 12.3       & -     \\ \cmidrule(l){2-8} 
                          & UL      & 2.0  & 44.8 & 41.8 & 51.9 \scriptsize{$\pm 1.12$}     & \textbf{12.3} \scriptsize{$\pm 0.42$} & 6.2   \\
                          & POP\rlap{\textit{\textsuperscript{$\flat$}}}    & 2.8  & 46.6 & 43.3 & 51.8 \scriptsize{$\pm 0.66$}     & 12.2 \scriptsize{$\pm 0.17$} & 6.4   \\
                          & POP     & 2.2  & 45.9 & 43.0 & \textbf{52.3} \scriptsize{$\pm 0.39$}     & \textbf{12.3} \scriptsize{$\pm 0.47$} & 6.2   \\ \bottomrule
\end{tabular}
\end{adjustbox}
}
\caption{\textbf{LM Performance Comparison.} The experimental results show the average accuracy over 9 classification tasks and the average F1 over 4 dialogue tasks. 
POP\textit{\textsuperscript{$\flat$}} is a method that utilizes $\mathcal{L}_{asc}$ and $\mathcal{L}_{ret}$ with hard labels, and POP employs $\mathcal{L}_{asc}$ and $\mathcal{L}_{ret}$ with soft labels. The best results are \textbf{bolded}.}
\label{maintable}
\end{table*}

\subsection{Evaluation Tasks}
Although POP is focused on unlearning a specific sequence of tokens, it is vital that the model performs well in all settings. 
Therefore, to ensure that the model is still capable of its original language modeling abilities post-unlearning, we evaluate the model on commonsense reasoning (Winogrande~\cite{sakaguchi2021winogrande} and COPA~\cite{gordon2012semeval}), linguistic reasoning (Hellaswag~\cite{zellers2019hellaswag} and Lambada~\cite{paperno2016lambada}), and scientific reasoning (ARC-Easy~\cite{clark2018think}, ARC-Challenge~\cite{clark2018think}, Piqa~\cite{bisk2020piqa}, MathQA~\cite{amini2019mathqa} PubmedQA~\cite{jin2019pubmedqa}) tasks. 
We also evaluate the model on dialogue tasks (Blended Skill Talk~\cite{smith2020can}, Empathetic Dialogues~\cite{rashkin2018towards}, Wizard of Internet~\cite{komeili2021internet}, and Wizard of Wikipedia~\cite{dinan2018wizard}) to assess the generation capabilities of the model. 

\subsection{Forgetting Thresholds} 
\label{sec:forgetthresholds}
We utilize EL\textsubscript{10}, MA, and RMA to determine when to stop the unlearning process. 
More specifically, we consider a token sequence $\textbf{x}^F$ to be forgotten when all three unlearning metrics fall below the average value on token sequences of Pile's evaluation set that were not seen during the pretraining.
This setting was also utilized in~\citet{jang2023knowledgeunlearning}, where they utilized thresholds for EL\textsubscript{10} and MA.\footnote{The threshold values for GPT-Neo may differ from \citet{jang2023knowledgeunlearning}, as we chose to utilize the uncopyrighted version of the Pile corpus to practice ethical research. For more details, please refer to Appendix~\ref{appendix:uncopyrighted_pile}.}
Table~\ref{forgetting_threshold} shows the threshold values for each metric, and the detailed process for calculating the thresholds can be found in Appendix~\ref{appendix:unlearning_criteria}.

\definecolor{blue_}{rgb}{0.122, 0.467, 0.706}
\definecolor{bluedark_}{rgb}{0.0976, 0.3736, 0.5648}

\definecolor{orange_}{rgb}{1.0, 0.498, 0.055}
\definecolor{orangedark_}{rgb}{0.8, 0.3984, 0.044}

\definecolor{purple_}{rgb}{0.173, 0.627, 0.173}
\definecolor{purpledark_}{rgb}{0.1384, 0.5016, 0.1384}

\definecolor{ulorange}{rgb}{0.996, 0.380, 0}
\definecolor{ulblue}{rgb}{0.392, 0.561, 1}
\definecolor{ulpink}{rgb}{0.863, 0.149, 0.498}

\definecolor{brickred}{rgb}{0.675, 0.243, 0.153}
\definecolor{bittersweet}{rgb}{0.722, 0.333, 0.169}
\definecolor{red_}{rgb}{0.914, 0.224, 0.212}

\begin{figure*}[h]
\vspace{-1cm}
\centering
  \begin{subfigure}{0.4\textwidth}
  \centering
  % 첫 번째 tikzpicture
    \scalebox{1.2}[0.9]{
    \begin{tikzpicture}[scale=0.6]
    \begin{axis}[
      xmin=0, xmax=10,
      ymin=30, ymax=55,
      xtick={0, 1, 2, 3, 4, 5, 6, 7, 8, 9, 10},
      ytick = {35, 40, 45, 50},
      ymajorgrids=true,
      major x tick style = transparent,
      grid style=dashed,
      legend entries={POP,UL},
      legend style={at={(1,1)},anchor=north east},
      xlabel = {\# of Batches Sequentially Forgotten},
      ylabel = {Avg. Classification Accuracy}
    ]
    \addplot[color=blue_, mark=diamond*, mark size=4pt, densely dashdotted] coordinates {(0, 53.81)(1, 52.16)(2, 49.93)(3, 47.13)(4, 47.13)(5, 47.13)(6, 44.21)(7, 44.21)(8, 44.21)(9, 44.21)(10, 44.21)}; %\addlegendentry{POP-2.7B}
    \addplot[color=red_, mark=o, mark size=3pt] coordinates {(0, 53.81)(1, 37.03)(2, 35.77)(3, 35.77)(4, 35.77)(5, 35.77)(6, 35.77)(7, 35.77)(8, 35.77)(9, 35.77)(10, 35.77)}; %\addlegendentry{UL-2.7B}  

    \end{axis}
    \end{tikzpicture}
    }
    \caption{Average accuracy for 9 Classification tasks.}
    \label{fig:1}
  \end{subfigure}
  \begin{subfigure}{0.4\textwidth}
  \centering
  % 첫 번째 tikzpicture
    \scalebox{1.2}[0.9]{
    \begin{tikzpicture}[scale=0.6]
    \begin{axis}[
      xmin=0, xmax=10,
      ymin=0, ymax=15,
      xtick={0, 1, 2, 3, 4, 5, 6, 7, 8, 9, 10},
      ytick = {3, 6, 9, 12},
      ymajorgrids=true,
      major x tick style = transparent,
      grid style=dashed,
      legend entries={POP,UL},
      legend style={at={(1,1)},anchor=north east},
      xlabel = {\# of Batches Sequentially Forgotten},
      ylabel = {Avg. Dialogue F1 score}
    ]
    \addplot[color=blue_, mark=diamond*, mark size=4pt, densely dashdotted] coordinates {(0, 13.69)(1, 13.29)(2, 11.05)(3, 8.97)(4, 8.97)(5, 8.97)(6, 6.55)(7, 6.55)(8, 6.55)(9, 6.55)(10, 6.55)}; %\addlegendentry{POP-2.7B}
    \addplot[color=red_, mark=o, mark size=3pt] coordinates {(0, 13.69)(1, 1.23)(2, 0.47)(3, 0.47)(4, 0.47)(5, 0.47)(6, 0.47)(7, 0.47)(8, 0.47)(9, 0.47)(10, 0.47)}; %\addlegendentry{UL-2.7B}  

    \end{axis}
    \end{tikzpicture}
    }
    
    % \caption{Neo-2.7B}
    \caption{Average F1 score for 4 Dialogue tasks.}
    \label{fig:1}
  \end{subfigure}
  \caption{\textbf{Sequential Unlearning Results.} We simulate a more likely scenario of complying to numerous unlearning requests with sequential unlearning experiments. The experiments were performed on the OPT 2.7B model, and the x-axis denotes the number of batches sequentially unlearned, with each batch containing 32 target sequences. The full results for all LMs tested are available in Appendix \ref{appendix:sequentialresults}.}
  \label{fig:sequential}
\end{figure*}
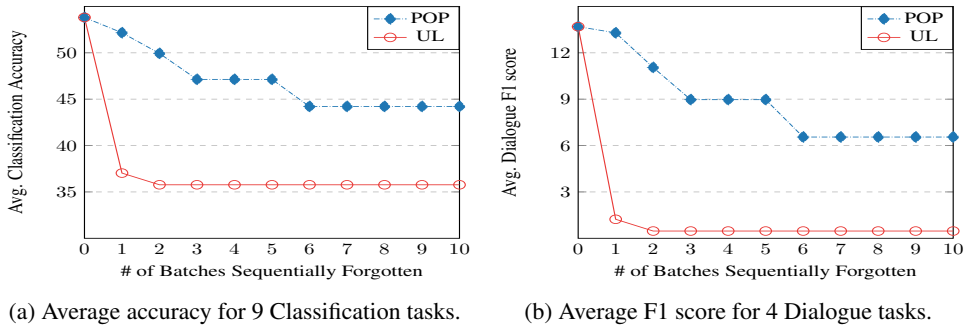

\section{Results and Analyses}
\subsection{Main Results}
We perform unlearning with 5 different random datasets of 32 target sequences, and report the averaged results for various OPT and GPT-Neo models in Table~\ref{maintable}. 
Individual results can be found in Appendix~\ref{appendix:individual_runs}.
Unlearning is performed until the model reaches the forgetting thresholds of all three metrics. 
The thresholds can be found in Table~\ref{forgetting_threshold}.
Here are our observations:

\noindent\textbf{(1)} Deduplicating the pretraining corpora can reduce the privacy risks, as OPT LMs show much smaller EL\textsubscript{10}, MA, and RMA values compared to the corresponding GPT-Neo models. 
However, deduplicating the corpora alone is not a valid unlearning solution, as the inherent privacy risk represented by EL\textsubscript{10}, MA, and RMA values are not significantly lower than that of GPT-Neo. 

\noindent\textbf{(2)} UL reaches the threshold much faster than the other two methods, demonstrated by the lower number of epochs required to reach the forgetting threshold. 
This is quite intuitive, as it only utilizes a single gradient ascent term, while the other two methods employ additional loss terms.
% in the gradient update process. 

\noindent\textbf{(3)} The actual EL\textsubscript{10}, MA, and RMA values for each model do not follow any pattern; that is, lower values do not necessarily indicate better performance.
Instead, they serve as a stopping threshold to confirm the completion of unlearning target tokens.

\noindent\textbf{(4)} UL performs the worst in both LMs for 9 classification and 4 dialogue benchmarks, showing degradation from the initial performance. 
This is even more evident in the OPT models, where the drop in performance is significant for dialogue tasks, potentially showing catastrophic forgetting. 
POP demonstrates the least amount of degradation, representing a remarkable retention of general language modeling capabilities. 

\noindent\textbf{(5)} UL demonstrates the largest variance in almost all benchmarks, which undermines its reliability and accentuates its dependence on the target token sequence to be unlearned. 

\noindent\textbf{(6)} We believe that the deduplication of Pile corpus on OPT models, along with the inclusion of other corpus in the training data, contributed to the extreme degradation in UL for OPT models. 
As GPT-Neo is trained solely on the Pile corpus, the duplicate instances might have contributed to the retention of LM performance after unlearning with UL. 
As most LMs include a wide range of corpora in their training sets, we believe that this further proves the strength of POP in demonstrating optimal unlearning \textit{and} retention of LM performance. 

\noindent\textbf{(7)} Although POP\rlap{\textit{\textsuperscript{$\flat$}}}  outperforms UL in most benchmarks, it fails to match the performance of POP.
This highlights the essential role of introducing the probability distribution over the vocabulary of the pretrained LM within $\mathcal{L}_{ret}$.
%The goal of sequence unlearning methods is to unlearn token sequences without significant degradation in language modeling performance. 

\begin{figure*}[h!]
\centering
\vspace{-1cm}
  \begin{subfigure}{0.5\textwidth}
    \centering
    \includegraphics[width=1.0\linewidth]{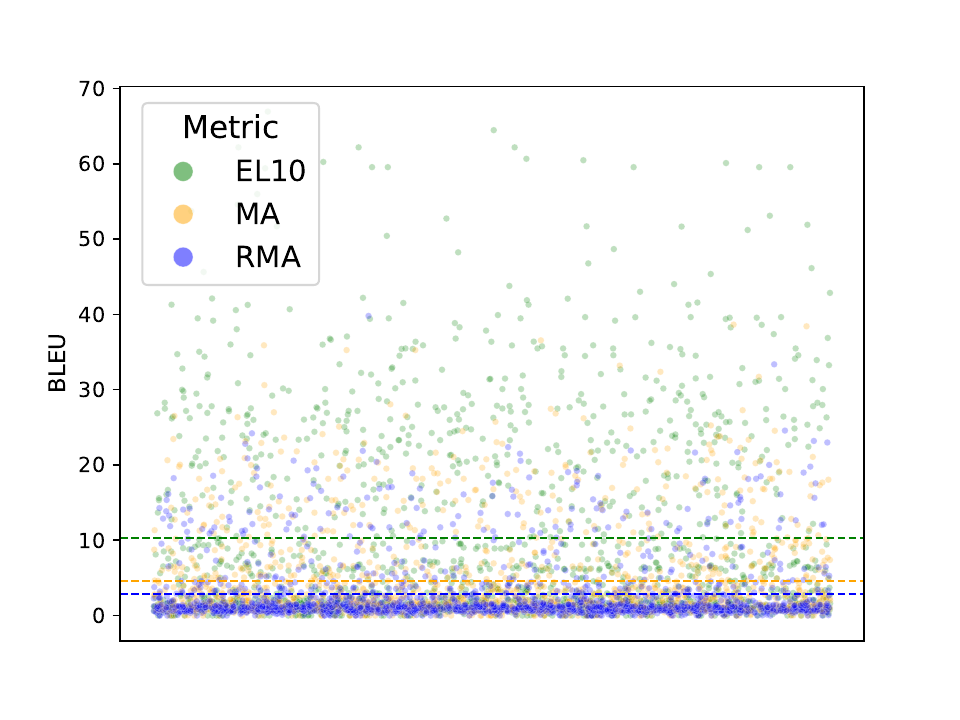}
    % \caption{BLEU score}
  \end{subfigure}%
  \begin{subfigure}{0.5\textwidth}
    \centering
    \includegraphics[width=1.0\linewidth]{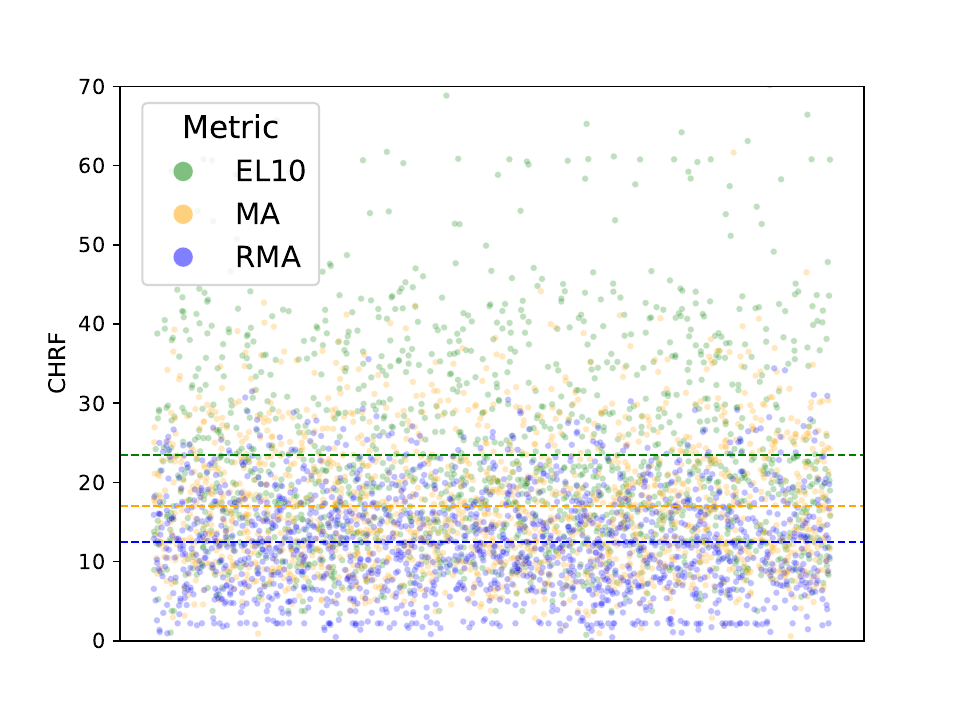}
    % \caption{CHRF score}
  \end{subfigure}  
  \caption{\textbf{Metric Comparison.} Models are unlearned until they reach the forgetting thresholds for each metric. After unlearning, we generate sequences with the resulting models, and compute BLEU and CHRF scores, where a lower score is favorable, as it indicates less overlap between the sequences. The dotted line represents the average scores for each metric. The data is spread out along the horizontal axis for visualization purposes.}
    \label{fig:metric_analysis}
\end{figure*}

\begin{figure*}[h!]
\centering
    \includegraphics[width=1.0\linewidth]{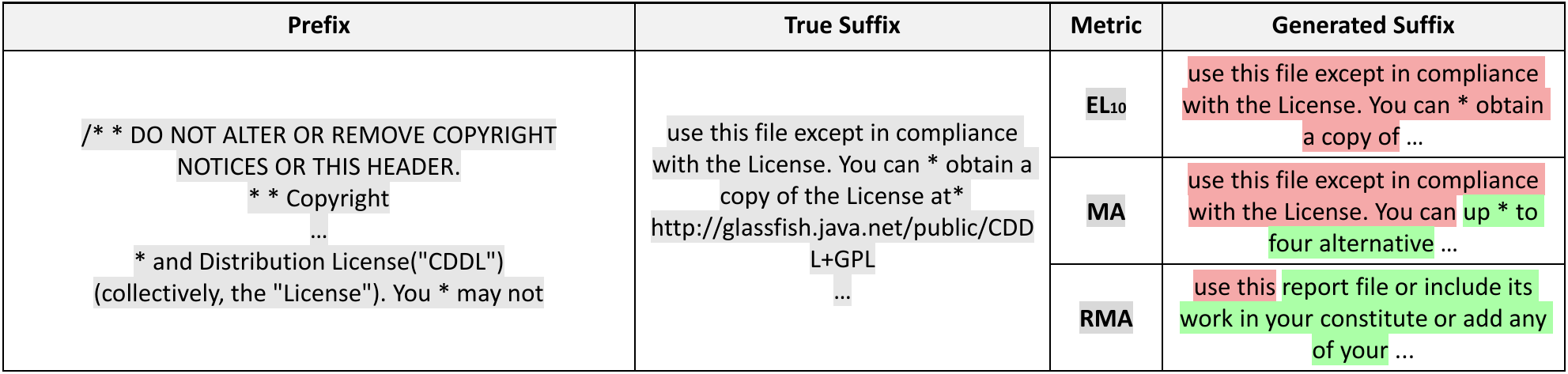}
    % \caption{CHRF score}
  \caption{Generated and True Suffixes for the given prefix. GPT-Neo LMs are unlearned with POP until the forgetting thresholds for each metric. \colorbox{red!30}{Red} indicates no unlearning, and \colorbox{green!30}{Green} indicates successful unlearning.}
    \label{fig:metric_example}
    
\end{figure*}

\subsection{Sequential Unlearning}
\label{sec:sequential_unlearning}

There are two ways to apply unlearning: batch unlearning and sequential unlearning. The results shown in Table~\ref{maintable} demonstrate batch unlearning results, in which all target sequences are unlearned at once. In sequential unlearning, target sequences are split into smaller batches, which are unlearned in succession.
Although batch unlearning is important to consider, sequential unlearning is a more likely real-world scenario, as unlearning requests will follow a sporadic pattern, requiring a more flexible solution. 

% To assess the practicality of POP, 
We sequentially unlearn 320 target sequences, split into 10 batches. 
Results for other models are available in Appendix \ref{appendix:sequentialresults}. 
As shown in Fig.~\ref{fig:sequential}, POP demonstrates better retention of performance in both classification and dialogue tasks compared to UL. 
After unlearning all 320 target sequences in 10 batches with UL, the performance of the OPT 2.7B model dropped over 18\% in average classification accuracy, and 13\% in average dialogue F1 score. 
The performance degradation in the dialogue task is extreme, as the average F1 score dropped to 0.47\%, demonstrating catastrophic forgetting of general LM capabilities. 
Furthermore, the performance in both sets of benchmarks reaches the minimum value after 2 batches, demonstrating the major flaw in UL.
POP, however, only demonstrates a moderate drop, demonstrating a decrease of 9.6\% for the average classification accuracy and 7.14\% for the average dialogue F1 score.
Fig.~\ref{fig:example} illustrates a qualitative example of the degradation in LM from UL.  
After the sequential unlearning of 10 batches with UL and POP, sequences are generated for a given prefix.
The generated sequence from the LM unlearned with the UL method demonstrates catastrophic degradation, while the LM unlearned with POP generates an acceptable response. 
UL is not a viable option, as repeated unlearning in succession with UL results in a catastrophic failure of LMs. 
On the other hand, POP successfully induces the LM to unlearn the target sequences \textit{and} does not significantly impact the LM performance. 

\subsection{Metric Analysis}
\label{sec:metric_analysis}
% \end{center}
We compare EL\textsubscript{10}, MA, and RMA by unlearning 3 separate GPT-Neo 2.7B models with POP, and stopping the unlearning process once they reach the forgetting thresholds for each metric. 
We generate 50 sequences for 1 target sequence using p-sampling with probabilities of p=0.9, 0.7, and 0.5, and use the first half of the sequence as a prefix to generate the second half as a suffix. 
Lastly, we compare the generated and the original sequences with BLEU~\cite{papineni2002bleu} and CHRF~\cite{popovic2015chrf}, where a lower score is favorable in the context of unlearning, as it indicates less overlap between the sequences.
As shown in Fig.~\ref{fig:metric_analysis}, models unlearned until the RMA threshold demonstrate the lowest BLEU and CHRF scores. 
This proves that in the context of unlearning, RMA provides the most privacy protection, as models that satisfy the RMA threshold are less likely to generate the original sequence. 
We also perform a qualitative analysis, which is shown on Fig. \ref{fig:metric_example}. 
It is clear that the model unlearned until the RMA threshold demonstrates the least amount of overlap between the sequences. 
Models unlearned until the EL\textsubscript{10} and MA thresholds, however, demonstrate some overlap in sequences, providing only partial unlearning. 
RMA provides the optimal privacy protection, demonstrating apt threshold for unlearning.  

\section{Conclusion}
In this paper, we propose POP, which effectively induces the LM to unlearn target token sequences without compromising its capabilities. 
We demonstrate the superior performance of POP in retaining LM performance on classification and dialogue benchmarks on two different LMs for three different sizes.
We also analyze a more likely scenario of complying to numerous unlearning request in succession with a sequential unlearning task, in which POP shows a much better retention of LM performance than previous work. 
Furthermore, we introduce RMA, a more stringent unlearning metric, and show how it can (1) better demonstrate the privacy risk of a LM, and (2) provide a stronger privacy protection when utilized to define an forgetting threshold.
We hope that researchers utilize the necessary privacy protection with POP to make LMs more viable for a wider range of tasks.

\section*{Limitations}
Despite the promising performance of POP, there are areas to expand upon our work.
Due to our experiments utilizing the Google Extraction benchmark, which is built on the Pile corpus, we inevitably experimented on GPT-Neo and OPT.
We leave applying POP to larger models as future work.
Due to the copyright issues, the forgetting threshold was determined based on the data samples chosen from the uncopyrighted Pile corpus, rather than original Pile corpus. 
It may result in a slight variance from previously reported the values.
Furthermore, as we mentioned in Section~\ref{datacuration}, we sampled the remaining data from the uncopyrighted Pile corpus, which does not include high-quality data, such as the book corpus.
This issue may have led to an inability to achieve further performance improvements.
Lastly, we were only able to simulate the real-world setting of sequential unlearning, which at times showed no changes to the results. 
This may have been due to the characteristics of the Training Data Extraction Challenge, which has overlap of data sources, such as code, which follow a very distinct style.
We leave the comprehensive analysis of sequential unlearning as future work to further investigate the application of sequence unlearning in LLMs.

\section*{Ethics Statement}
To promote transparency within the natural language community, many have promoted the move towards removing copyrighted content from LMs. Furthermore, as the goal of our research is to improve the LLM's privacy guarantees, we were encouraged to only utilize the uncopyrighted version of the Pile corpus. 
All experiments were conducted on English datasets, where we looked to induce unlearning of English sequences from publicly available LMs. Utilizing the method on non-English models is not verified. 
Lastly, resulting models post-unlearning may generate hallucinations, which is an unintended side effect of LMs, but also an inherent problem with LMs. 

\section*{Acknowledgements}
This work was supported by Institute for Information \& communications Technology Promotion(IITP) grant funded by the Korea government(MSIT) (No.RS-2019-II190075 Artificial Intelligence Graduate School Program(KAIST)) and the National Research Foundation of Korea (NRF) grant funded by the Korea government (MSIT) (No. NRF-2022R1A2B5B02001913) and Samsung Electronics Co., Ltd.

\bibliography{anthology,custom}

% \clearpage
\appendix
\section*{Additional Details for POP}

\section{Training Details}
We conduct the experiments with the learning rate at 5e-5 with constant scheduling, and both dropout and weight decay were set to 0. We set $\lambda=1$, the loss hyperparameter described in equation~\ref{eq:11}.  
We implement with Pytorch~\cite{pytorch} and Pytorch Lightning~\cite{pytorch_lightning}. 
We load GPT-Neo and OPT models (125M, 1.3B, 2.7B) from Hugging Face's Transformers~\cite{huggingface}. We utilize DeepSpeed ZeRO Stage 2 Offload and FusedAdam~\cite{deepspeed}, along with fp16 mixed precision~\cite{fp16}. 
The batch size is 8, and gradient accumulation is used to update all mini-batches simultaneously.
During each unlearning step, we use 32 retain data for training. 
We use NVIDIA RTX A6000 and 3090 GPUs; the unlearning process takes approximately 1 hour for the 125M model and around 3 hours for the 1.3B and 2.7B models.

\section{Uncopyrighted Pile Corpus}
\label{appendix:uncopyrighted_pile}
\begin{table}[h!]
\centering
\small{
\begin{tabular}{@{}cc|ccc@{}}
\toprule
                      & Size & EL\textsubscript{10} & MA   & RMA  \\ \midrule
\multirow{3}{*}{\raisebox{-2.0mm}{\hspace{3mm} Ours}} & 125M & 6.3  & 48.7 & 41.5 \\ \cmidrule(l){2-5} 
                      & 1.3B & 7.9  & 54.2 & 48.1 \\ \cmidrule(l){2-5} 
                      & 2.7B & 8.5  & 55.5 & 49.6 \\ \midrule \midrule
\multirow{3}{*}{\raisebox{-2.0mm}{\hspace{3mm}\citeauthor{jang2023knowledgeunlearning}}} & 125M & 5.0  & 29.9 & -    \\ \cmidrule(l){2-5} 
                      & 1.3B & 5.7  & 33.3 & -    \\ \cmidrule(l){2-5} 
                      & 2.7B & 5.5  & 34.0 & -    \\ \bottomrule
\end{tabular}
}
\caption{Threshold comparison for GPT-Neo}
\label{threshold_comparison}
\end{table}
\noindent The original Pile corpus~\cite{gao2020pile} is not available anymore due to copyright issues. 
To practice ethical research, we utilized the uncopyrighted Pile corpus\footnote{\url{https://huggingface.co/datasets/monology/pile-uncopyrighted}} and computed all thresholds in Appendix~\ref{appendix:unlearning_criteria}. 
The uncopyrighted version of the Pile corpus removes Books3, BookCorpus2, OpenSubtitles, YTSubtitles, and OWT2 from the original dataset, which is a significant portion of the dataset.
Although we utilized the same process in computing the thresholds as \citet{jang2023knowledgeunlearning}, the removal of copyrighted data impacted the threshold values. 
Table~\ref{threshold_comparison} shows the different threshold values for GPT-Neo.
Although this may have led to discrepancies between the performance of the UL method presented in \citet{jang2023knowledgeunlearning} and in our experiment for GPT-Neo model, we believe that the differences are minimal, and will not impact the relative performance of the methods.

\section{Measuring Forgetting Thresholds}
\label{appendix:unlearning_criteria}

\begin{table}[h]
\centering
\begin{tabular}{@{}c|c@{}}
\toprule
Domain            & Number of data \\ \midrule
Pile-CC           & 2739           \\
PubMed Central    & 1920           \\
ArXiv             & 1190           \\
Github            & 1010           \\
FreeLaw           & 820            \\
StackExchange     & 680            \\
USPTO Backgrounds & 490            \\
PubMed Abstracts  & 410            \\
Wikipedia (en)    & 200            \\
DM Mathematics    & 170            \\
EuroParl          & 100            \\
HackerNews        & 80             \\
Gutenberg (PG-19) & 60             \\
PhilPapers        & 50             \\
NIH ExPorter      & 40             \\
Ubuntu IRC        & 21             \\
Enron Emails      & 20             \\ \bottomrule
\end{tabular}
\caption{The number of data used for measuring the forgetting thresholds for each domain.}
\label{table:data used for measuring forgetting threshold}
\end{table}

\noindent For measuring the forgetting threshold, we used the uncopyrighted Pile corpus to conduct research ethically. 
We sampled 10,000 data through weighted sampling based on the domain distribution of the Pile corpus. 
Table \ref{table:data used for measuring forgetting threshold} shows the number of sampled data for each domain.
We measured the thresholds for EL\textsubscript{10}, MA, and RMA, and the results are presented in Table \ref{forgetting_threshold}.

\onecolumn
\raggedright
\section{Sequential Unlearning Results}
\label{appendix:sequentialresults}
\begin{figure*}[h!]
\centering
  \begin{subfigure}{0.4\textwidth}
  \centering
    \scalebox{1.2}[0.9]{

    }
    
    \caption{OPT-125M}
    \label{fig:1}
  \end{subfigure}
  \begin{subfigure}{0.4\textwidth}
  \centering
  % 첫 번째 tikzpicture
    %\resizebox{7.5cm}{6.5cm}{
    \scalebox{1.2}[0.9]{
    %
    }
    
    \caption{OPT-1.3B}
    \label{fig:1}
  \end{subfigure}
  
  \medskip
  \begin{subfigure}{0.4\textwidth}
  \centering
  % 첫 번째 tikzpicture
    %\resizebox{7.5cm}{6.5cm}{
    \scalebox{1.2}[0.9]{
    %
    }
    \caption{OPT-125M}
    \label{fig:1}
  \end{subfigure}
  \begin{subfigure}{0.4\textwidth}
  \centering
  % 첫 번째 tikzpicture
    %\resizebox{7.5cm}{6.5cm}{
    \scalebox{1.2}[0.9]{
    %
    }
    
    \caption{OPT-1.3B}
    \label{fig:1}
  \end{subfigure}
  
  \caption{The first row indicates the average accuracy for 9 classification tasks, and the second row shows the average F1 score for 4 Dialogue tasks for OPT models.}
  \label{fig:images}
\end{figure*}

\begin{figure*}[h!]
\centering
  \begin{subfigure}{0.3\textwidth}
  \centering
  % 첫 번째 tikzpicture
    %\resizebox{7.5cm}{6.5cm}{
    % \scalebox{1.2}[0.9]{
    %
    % }
    
    \caption{Neo-125M}
    \label{fig:1}
  \end{subfigure}
  \begin{subfigure}{0.3\textwidth}
  \centering
  % 첫 번째 tikzpicture
    %\resizebox{7.5cm}{6.5cm}{
    % \scalebox{1.2}[0.9]{
    %
    % }
    
    \caption{Neo-1.3B}
    \label{fig:1}
  \end{subfigure}
  \begin{subfigure}{0.3\textwidth}
  \centering
  % 첫 번째 tikzpicture
    %\resizebox{7.5cm}{6.5cm}{
    % \scalebox{1.2}[0.9]{
    %
    % }
    
    \caption{Neo-2.7B}
    \label{fig:1}
  \end{subfigure}
  \medskip
  \begin{subfigure}{0.3\textwidth}
  \centering
  % 첫 번째 tikzpicture
    %\resizebox{7.5cm}{6.5cm}{
    % \scalebox{1.2}[0.9]{
    %
    % }
    \caption{Neo-125M}
    \label{fig:1}
  \end{subfigure}
  \begin{subfigure}{0.3\textwidth}
  \centering
  % 첫 번째 tikzpicture
    %\resizebox{7.5cm}{6.5cm}{
    % \scalebox{1.2}[0.9]{
    %

    \caption{Neo-1.3B}
    \label{fig:1}
  \end{subfigure}
  \begin{subfigure}{0.3\textwidth}
  \centering
  % 첫 번째 tikzpicture
    %\resizebox{7.5cm}{6.5cm}{
    % \scalebox{1.2}[0.9]{
    %
    % }
    
    \caption{Neo-125M}
    \label{fig:1}
  \end{subfigure}
  
  \caption{The first row indicates the average accuracy for 9 classification tasks, and the second row shows the average F1 score for 4 Dialogue tasks for GPT-Neo models.}
  \label{fig:images}
\end{figure*}
\clearpage

\section{Individual Runs}
\label{appendix:individual_runs}

\begin{table}[h!]
\scalebox{0.6}{
\begin{tabular}{c|ccc|ccc|ccccccccc|cccc}
\hline
                                  & Method                & Metric & Epoch & EL\textsubscript{10} & MA   & RMA  & Lamba. & Piqa & Hella. & ARC-E & ARC-C & Copa & Wino. & MathQ & PubQ & Wiki & Inter. & Empa. & Blend. \\ \hline
\multirow{10}{*}{Forgetting set0} & Pretrained            & -      & -     & 9.1  & 58.1 & 45.3 & 38.9   & 62.0 & 28.5   & 45.2  & 20.7  & 66.0 & 53.2  & 21.8  & 47.4 & 11.1 & 12.5   & 9.2   & 10.4   \\ \cline{2-20} 
                                  & \multirow{3}{*}{UL}   & EL\textsubscript{10}   & 9     & 4.2  & 41.4 & 37.6 & 32.0   & 60.1 & 27.7   & 35.6  & 19.0  & 65.0 & 51.3  & 21.4  & 36.4 & 10.2 & 11.8   & 9.7   & 10.4   \\
                                  &                       & MA    & 11    & 4.1  & 37.5 & 35.0 & 19.6   & 58.4 & 27.2   & 30.2  & 18.6  & 54.0 & 51.1  & 21.3  & 33.0 & 7.6  & 9.7    & 6.7   & 8.0    \\
                                  &                       & RMA    & 14    & 2.8  & 30.3 & 29.0 & 2.6    & 57.0 & 27.1   & 28.8  & 20.0  & 55.0 & 50.7  & 20.6  & 32.4 & 2.2  & 3.0    & 1.2   & 1.7    \\ \cline{2-20} 
                                  & \multirow{3}{*}{POP}  & EL\textsubscript{10}   & 11    & 4.2  & 36.0 & 34.3 & 37.1   & 62.1 & 28.3   & 46.0  & 20.0  & 65.0 & 52.5  & 22.0  & 47.2 & 11.1 & 11.6   & 9.1   & 10.1   \\
                                  &                       & MA    & 14    & 5.0  & 39.6 & 37.3 & 39.0   & 62.3 & 28.2   & 45.2  & 20.7  & 67.0 & 52.9  & 21.6  & 50.4 & 11.5 & 12.0   & 9.6   & 10.9   \\
                                  &                       & RMA    & 14    & 2.8  & 31.8 & 30.9 & 39.5   & 62.4 & 28.3   & 44.4  & 21.7  & 68.0 & 52.6  & 21.7  & 53.0 & 9.8  & 11.2   & 7.5   & 9.4    \\
\hline
\end{tabular}
}
\caption{All of the individual runs for OPT 125M. 
The \textbf{Metric} column indicates the checkpoint at which the given metric reaches the pre-defined threshold. In Table~\ref{maintable}, we reported the result when all metrics are satisfied with each threshold.
}
\end{table}

% \usepackage{multirow}
\begin{table}[h!]
\scalebox{0.6}{
\begin{tabular}{c|ccc|ccc|ccccccccc|cccc}
\hline
                                  & Method                & Metric & Epoch & EL\textsubscript{10} & MA   & RMA  & Lamba. & Piqa & Hella. & ARC-E & ARC-C & Copa & Wino. & MathQ & PubQ & Wiki & Inter. & Empa. & Blend. \\ \hline
\multirow{10}{*}{Forgetting set0} & Pretrained            & -      & -     & 29.9 & 70.9 & 64.3 & 58.9   & 71.6 & 39.7   & 55.6  & 24.1  & 76   & 56.7  & 23.2  & 57.8 & 13.0 & 13.7   & 12.6  & 13.8   \\ \cline{2-20} 
                                  & \multirow{3}{*}{UL}   & EL\textsubscript{10}   & 5     & 4.3  & 57.7 & 53.1 & 58.2   & 71.2 & 40.0   & 52.6  & 23.7  & 75.0 & 56.8  & 23.4  & 58.0 & 13.0 & 14.6   & 12.6  & 13.9   \\
                                  &                       & MA    & 5     & 4.0  & 39.3 & 36.8 & 20.5   & 58.6 & 32.6   & 30.9  & 23.4  & 64.0 & 51.1  & 21.4  & 47.4 & 4.1  & 4.8    & 2.4   & 4.8    \\
                                  &                       & RMA    & 6     & 4.0  & 39.3 & 36.8 & 20.5   & 58.6 & 32.6   & 30.9  & 23.4  & 64.0 & 51.1  & 21.4  & 47.4 & 4.1  & 4.8    & 2.4   & 4.8    \\ \cline{2-20} 
                                  & \multirow{3}{*}{POP}  & EL\textsubscript{10}   & 6     & 5.7  & 55.8 & 53.1 & 60.8   & 71.2 & 39.7   & 52.7  & 25.4  & 76.0 & 56.3  & 23.5  & 58.0 & 13.6 & 13.9   & 12.7  & 13.7   \\
                                  &                       & MA    & 6     & 3.7  & 37.9 & 35.9 & 58.2   & 70.8 & 38.4   & 52.4  & 24.4  & 75.0 & 56.1  & 22.7  & 58.0 & 11.7 & 13.5   & 12.3  & 12.8   \\
                                  &                       & RMA    & 7     & 3.7  & 37.9 & 35.9 & 58.2   & 70.8 & 38.4   & 52.4  & 24.4  & 75.0 & 56.1  & 22.7  & 58.0 & 11.7 & 13.5   & 12.3  & 12.8   \\
\hline
\end{tabular}
}
\caption{All of the individual runs for OPT 1.3B. 
The \textbf{Metric} column indicates the checkpoint at which the given metric reaches the pre-defined threshold. In Table~\ref{maintable}, we reported the result when all metrics are satisfied with each threshold.
}
\end{table}

% \usepackage{multirow}
\begin{table}[h!]
\scalebox{0.6}{
\begin{tabular}{c|ccc|ccc|ccccccccc|cccc}
\hline
                                  & Method                & Metric & Epoch & EL\textsubscript{10} & MA   & RMA  & Lamba. & Piqa & Hella. & ARC-E & ARC-C & Copa & Wino. & MathQ & PubQ & Wiki & Inter. & Empa. & Blend. \\ \hline
\multirow{10}{*}{Forgetting set0} & Pretrained            &        &       & 32.2 & 72.6 & 66.3 & 64.4   & 74.3 & 43.5   & 56.8  & 27.1  & 78.0 & 59.1  & 23.0  & 58.2 & 13.4 & 14.7   & 13.0  & 13.6   \\ \cline{2-20} 
                                  & \multirow{3}{*}{UL}   & EL\textsubscript{10}   & 6     & 5.6  & 60.8 & 57.3 & 63.6   & 72.6 & 42.4   & 50.4  & 28.1  & 72.0 & 58.9  & 23.0  & 57.0 & 12.9 & 15.1   & 12.3  & 14.1   \\
                                  &                       & MA    & 8     & 3.5  & 40.6 & 39.1 & 20.2   & 62.0 & 34.3   & 35.1  & 23.7  & 59.0 & 51.6  & 21.8  & 53.2 & 3.3  & 3.9    & 3.4   & 4.6    \\
                                  &                       & RMA    & 8     & 3.5  & 40.6 & 39.1 & 20.2   & 62.0 & 34.3   & 35.1  & 23.7  & 59.0 & 51.6  & 21.8  & 53.2 & 3.3  & 3.9    & 3.4   & 4.6    \\ \cline{2-20} 
                                  & \multirow{3}{*}{POP}  & EL\textsubscript{10}   & 6     & 6.0  & 53.8 & 51.9 & 64.7   & 73.3 & 42.5   & 56.3  & 29.2  & 74.0 & 59.0  & 22.4  & 57.6 & 13.7 & 14.6   & 13.6  & 13.8   \\
                                  &                       & MA    & 8     & 4.3  & 44.0 & 42.6 & 62.5   & 73.0 & 41.9   & 55.4  & 28.1  & 73.0 & 58.6  & 23.0  & 57.6 & 13.2 & 14.4   & 13.2  & 13.6   \\
                                  &                       & RMA    & 8     & 4.7  & 36.2 & 35.6 & 60.5   & 72.7 & 41.3   & 55.6  & 27.1  & 73.0 & 58.2  & 23.2  & 57.6 & 12.8 & 14.3   & 13.0  & 13.3   \\
\hline
\end{tabular}
}
\caption{All of the individual runs for OPT 2.7B. 
The \textbf{Metric} column indicates the checkpoint at which the given metric reaches the pre-defined threshold. In Table~\ref{maintable}, we reported the result when all metrics are satisfied with each threshold.
}
\end{table}

\begin{table*}[h!]
\centering
\scalebox{0.6}{
\begin{tabular}{c|ccc|ccc|ccccccccc|cccc}
\hline
                                                   & Method                & Metric & Epoch & EL\textsubscript{10}   & MA  & RMA  & Lamba. & Piqa & Hella. & ARC-E & ARC-C & Copa & Wino. & MathQ & PubQ & Wiki & Inter. & Empa. & Blend. \\ \hline
\multicolumn{1}{c|}{\multirow{10}{*}{Forgetting set0}} & Pretrained                 & -    & -     & 41.5 & 80.5 & 74.0 & 37.6   & 63.4 & 28.2   & 45.7  & 22.0  & 63.0 & 51.5  & 22.5  & 57.6 & 10.5 & 11.3   & 8.4   & 9.7    \\ \cline{2-20} 
\multicolumn{1}{c|}{}                              & \multirow{3}{*}{UL}   & EL\textsubscript{10} & 9     & 5.0  & 56.5 & 51.9 & 27.5   & 61.2 & 28.0   & 44.3  & 23.1  & 62.0 & 51.1  & 22.1  & 57.4 & 9.6  & 10.3   & 8.1   & 8.3    \\
\multicolumn{1}{c|}{}                              &                       & MA  & 11    & 3.1  & 47.5 & 43.3 & 20.5   & 61.0 & 27.8   & 42.9  & 24.1  & 61.0 & 51.1  & 21.9  & 54.0 & 8.9  & 9.5    & 7.7   & 7.8    \\
\multicolumn{1}{c|}{}                              &                       & RMA  & 14    & 2.9  & 42.7 & 38.5 & 17.8   & 60.7 & 27.8   & 41.8  & 23.7  & 61.0 & 51.2  & 21.9  & 53.4 & 8.4  & 9.3    & 7.4   & 7.5    \\ \cline{2-20} 
\multicolumn{1}{c|}{}                              & \multirow{3}{*}{POP}  & EL\textsubscript{10} & 11    & 5.3  & 57.2 & 52.2 & 36.4   & 63.4 & 28.1   & 43.9  & 20.7  & 60.0 & 52.1  & 22.5  & 57.6 & 10.9 & 11.4   & 9.0   & 9.8    \\
\multicolumn{1}{c|}{}                              &                       & MA  & 14    & 1.9  & 44.9 & 40.1 & 37.8   & 63.3 & 28.0   & 43.2  & 20.0  & 60.0 & 51.9  & 22.4  & 57.4 & 10.9 & 11.7   & 9.0   & 9.7    \\
\multicolumn{1}{c|}{}                              &                       & RMA  & 14    & 1.9  & 44.9 & 40.1 & 37.8   & 63.3 & 28.0   & 43.2  & 20.0  & 60.0 & 51.9  & 22.4  & 57.4 & 10.9 & 11.7   & 9.0   & 9.7    \\
\hline
\end{tabular}
}
\caption{All of the individual runs for GPT-Neo 125M. 
The \textbf{Metric} column indicates the checkpoint at which the given metric reaches the pre-defined threshold. In Table~\ref{maintable}, we reported the result when all metrics are satisfied with each threshold.
}
\end{table*}

% \usepackage{multirow}
\begin{table*}[h!]
\scalebox{0.6}{
\begin{tabular}{c|ccc|ccc|ccccccccc|cccc}
\hline
                                                   & Method                & Metric & Epoch & EL\textsubscript{10}   & MA  & RMA  & Lamba. & Piqa & Hella. & ARC-E & ARC-C & Copa & Wino. & MathQ & PubQ & Wiki & Inter. & Empa. & Blend. \\ \hline
\multicolumn{1}{c|}{\multirow{10}{*}{Forgetting set0}} & Pretrained                 & -    & -     & 66.1 & 91.8 & 88.1 & 57.2   & 70.4 & 37.0   & 56.6  & 25.8  & 70.0 & 54.6  & 21.9  & 53.6 & 12.7 & 13.8   & 10.5  & 12.1   \\ \cline{2-20} 
\multicolumn{1}{c|}{}                              & \multirow{3}{*}{UL}   & EL\textsubscript{10} & 5     & 3.6  & 53.1 & 48.4 & 65.6   & 70.4 & 37.3   & 56.8  & 26.1  & 68.0 & 56.2  & 21.9  & 55.4 & 12.4 & 12.6   & 9.9   & 10.8   \\
\multicolumn{1}{c|}{}                              &                       & MA  & 5     & 3.6  & 53.1 & 48.4 & 65.6   & 70.4 & 37.3   & 56.8  & 26.1  & 68.0 & 56.2  & 21.9  & 55.4 & 12.4 & 12.6   & 9.9   & 10.8   \\
\multicolumn{1}{c|}{}                              &                       & RMA  & 6     & 2.1  & 42.1 & 37.0 & 67.1   & 70.2 & 37.3   & 55.6  & 26.1  & 68.0 & 56.7  & 22.0  & 55.6 & 11.3 & 11.4   & 9.1   & 10.1   \\ \cline{2-20} 
\multicolumn{1}{c|}{}                              & \multirow{3}{*}{POP}  & EL\textsubscript{10} & 6     & 4.1  & 53.7 & 49.4 & 56.8   & 71.0 & 37.3   & 56.1  & 26.1  & 70.0 & 54.8  & 21.7  & 53.6 & 12.6 & 14.0   & 10.5  & 12.2   \\
\multicolumn{1}{c|}{}                              &                       & MA  & 6     & 4.1  & 53.7 & 49.4 & 56.8   & 71.0 & 37.3   & 56.1  & 26.1  & 70.0 & 54.8  & 21.7  & 53.6 & 12.6 & 14.0   & 10.5  & 12.2   \\
\multicolumn{1}{c|}{}                              &                       & RMA  & 7     & 2.2  & 46.4 & 42.1 & 55.6   & 70.7 & 37.4   & 56.1  & 26.8  & 71.0 & 55.3  & 21.9  & 54.4 & 12.4 & 13.8   & 10.6  & 12.3   \\
\hline
\end{tabular}
}
\caption{All of the individual runs for GPT-Neo 1.3B. 
The \textbf{Metric} column indicates the checkpoint at which the given metric reaches the pre-defined threshold. In Table~\ref{maintable}, we reported the result when all metrics are satisfied with each threshold.
}
\end{table*}

\begin{table*}[h!]
\scalebox{0.6}{
\begin{tabular}{c|ccc|ccc|ccccccccc|cccc}
\hline
                                                   & Method                & Metric & Epoch & EL\textsubscript{10}   & MA  & RMA  & Lamba. & Piqa & Hella. & ARC-E & ARC-C & Copa & Wino. & MathQ & PubQ & Wiki & Inter. & Empa. & Blend. \\ \hline
\multicolumn{1}{c|}{\multirow{10}{*}{Forgetting set0}} & Pretrained                 & -    & -     & 68.3 & 92.7 & 90.0 & 62.3   & 73.0 & 40.7   & 59.8  & 25.4  & 74.0 & 56.1  & 21.4  & 57.0 & 12.4 & 13.7   & 10.9  & 12.4   \\ \cline{2-20} 
\multicolumn{1}{c|}{}                              & \multirow{3}{*}{UL}   & EL\textsubscript{10} & 6     & 3.9  & 56.0 & 53.3 & 64.8   & 73.2 & 41.1   & 58.6  & 29.8  & 72.0 & 55.2  & 21.3  & 57.4 & 13.1 & 13.8   & 11.7  & 12.5   \\
\multicolumn{1}{c|}{}                              &                       & MA  & 8     & 2.5  & 37.2 & 34.9 & 65.5   & 73.2 & 41.3   & 57.7  & 29.2  & 72.0 & 55.5  & 21.4  & 57.4 & 13.2 & 13.7   & 11.3  & 12.3   \\
\multicolumn{1}{c|}{}                              &                       & RMA  & 8     & 2.5  & 37.2 & 34.9 & 65.5   & 73.2 & 41.3   & 57.7  & 29.2  & 72.0 & 55.5  & 21.4  & 57.4 & 13.2 & 13.7   & 11.3  & 12.3   \\ \cline{2-20} 
\multicolumn{1}{c|}{}                              & \multirow{3}{*}{POP}  & EL\textsubscript{10} & 6     & 6.7  & 61.1 & 58.9 & 63.5   & 73.1 & 41.3   & 58.9  & 28.1  & 73.0 & 54.9  & 21.4  & 57.4 & 12.3 & 13.3   & 10.4  & 12.0   \\
\multicolumn{1}{c|}{}                              &                       & MA  & 8     & 2.3  & 44.7 & 42.8 & 62.6   & 73.2 & 41.2   & 58.6  & 28.5  & 72.0 & 54.5  & 21.4  & 57.4 & 12.6 & 13.4   & 10.6  & 12.4   \\
\multicolumn{1}{c|}{}                              &                       & RMA  & 8     & 2.3  & 44.7 & 42.8 & 62.6   & 73.2 & 41.2   & 58.6  & 28.5  & 72.0 & 54.5  & 21.4  & 57.4 & 12.6 & 13.4   & 10.6  & 12.4   \\
\hline
\end{tabular}
}
\caption{All of the individual runs for GPT-Neo 2.7B. 
The \textbf{Metric} column indicates the checkpoint at which the given metric reaches the pre-defined threshold. In Table~\ref{maintable}, we reported the result when all metrics are satisfied with each threshold.
}
\end{table*}

\end{document}